\pdfoutput=1

\documentclass[11pt]{article}

\usepackage{acl2023}

\usepackage{times}
\usepackage{latexsym}
\usepackage{amsmath}
\usepackage{graphicx}
\usepackage{subcaption}
\usepackage{booktabs}
\usepackage{hyperref}
\usepackage{cleveref}
\usepackage{booktabs}
\usepackage{dblfloatfix}
\usepackage{multirow}

\usepackage[T1]{fontenc}
\usepackage{xcolor}
\usepackage{colortbl}

\usepackage[utf8]{inputenc}
\usepackage{float}
\usepackage{caption}
\usepackage{algorithm}
\usepackage{algorithmic}

\usepackage{microtype}
\usepackage{inconsolata}

\title{PolyPrompt: Automating Knowledge Extraction from Multilingual Language Models with Dynamic Prompt Generation}

\author{
  Nathan Roll\\
  Stanford University\\
  \texttt{nroll@stanford.edu}
}

\begin{document}
\maketitle

\begin{abstract}
Large language models (LLMs) showcase increasingly impressive English benchmark scores, however their performance profiles remain inconsistent across multilingual settings. To address this gap, we introduce \textbf{PolyPrompt}, a novel, parameter-efficient framework for enhancing the multilingual capabilities of LLMs. Our method learns a set of trigger tokens for each language through a gradient-based search, identifying the input query's language and selecting the corresponding trigger tokens which are prepended to the prompt during inference. We perform experiments on two $\sim$1 billion parameter models, with evaluations on the global MMLU benchmark across fifteen typologically and resource diverse languages, demonstrating accuracy gains of 3.7\%-19.9\% compared to naive and translation-pipeline baselines.
\end{abstract}
\section{Introduction}
Large language models (LLMs) trained on multilingual data offer a clear value for non-English speakers. However, these models exhibit a significant performance degradation in non-English languages. This bias arises from several interconnected factors: (1) the relative prevalence of English in training/fine-tuning corpora \cite{dodge2021documenting, gao2020pile}; (2) the underrepresentation of diverse linguistic perspectives in AI research and development \cite{ahmed2023state}; and (3) the historical dominance of English-language benchmarks in evaluating model capabilities \cite{lin2021truthfulqa, srivastava2022beyond}.

Prompt engineering, crafting effective inputs for LLMs, is crucial for extracting the best results from such systems \cite{liu2021pre}. While traditionally manual, automated prompt generation, or \textit{autoprompting} \cite{shin2020autoprompt, wallace2019universal, guo2021textual}, offers a scalable and less labor-intensive alternative. Autoprompting uses gradient-based search to discover ``prompts'' that optimize performance, offering computational efficiency over methods which modify model weights themselves. However, existing autoprompting methods are often applied in monolingual settings \citep{shin2020autoprompt} or rely on static, translated prompts, failing to account for language-specific LLM behaviors.

\begin{figure}
    \centering
    \includegraphics[width=1.05\linewidth]{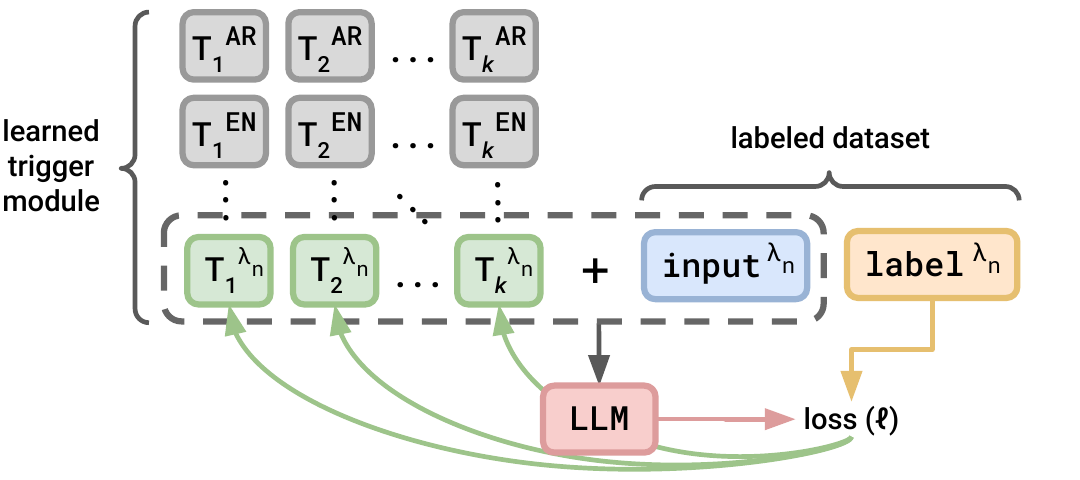}
    \caption{PolyPrompt learns a set of trigger tokens for each language, using a labeled dataset, to improve multilingual LLM performance.}
    \label{fig:polyprompt-diagram}
\end{figure}

We introduce \textbf{PolyPrompt}, a novel autoprompting framework designed to mitigate language-based disparities in pre-trained model performance. Unlike static or translated prompts, PolyPrompt learns and dynamically applies \emph{language-specific trigger tokens}. Our method detects the input query's language and activates the corresponding optimized trigger tokens, moving beyond simple translation-based approaches (see \cref{fig:polyprompt-diagram}). By dynamically adapting prompts to the input language, our experiments reveal that PolyPrompt unlocks the latent multilingual capabilities of LLMs in all tested languages\footnote{See \Cref{sec:appendix-languages} for language table.}: \texttt{am, ar, cs, de, en, es, fa, fr, hi, it, ja, ko, nl, sw, zh}.

The contributions of this work include: (1) PolyPrompt, a novel dynamic autoprompting framework, (2) a code implementation to automatically select and apply language-specific trigger tokens based on the detected language, and (3) a comprehensive evaluation on the global MMLU \citep{singh2024globalmmluunderstandingaddressing} benchmark.

\section{Methods}
\label{sec:polyprompt-method}
\subsection{Data}
We evaluate PolyPrompt on the Global MMLU dataset \citep{singh2024globalmmluunderstandingaddressing}, based on \citet{hendrycks2020measuring}. MMLU is a challenging benchmark for evaluating multilingual reasoning, making it suitable for assessing PolyPrompt's effectiveness in mitigating language bias. We use the test subsets in fifteen languages. Each subset contains 14,042 professionally translated multiple-choice questions. For each language, we randomly split the dataset, reserving 20\% for evaluation and using the remaining 80\% for learning trigger embeddings.

\subsection{Models}
We employ Llama 3 1b base and instruct models \citep{meta2024llama3-2-1b} to demonstrate the parameter efficiency and effectiveness of PolyPrompt.  We denote the pre-trained language model as $f_\theta$ and freeze all its parameters during training. Only the \emph{trigger embeddings} are updated, ensuring parameter efficiency and modularity in learning language-specific prompts.

\subsection{Training Procedure}

\paragraph{Trigger Embeddings.}
PolyPrompt learns a set of language-specific \emph{trigger embeddings}. For each language $\lambda \in \{\lambda_1, \ldots, \lambda_n\}$, we maintain a matrix of $k$ learnable embeddings, $T^\lambda \in {R}^{k \times d}$, where $k$ is the number of trigger tokens and $d$ is the embedding dimension of the language model. These embeddings are initialized randomly and optimized for each language independently. \footnote{See \cref{sec:appendix-notes} for implementation details, including hyperparameters.}

\paragraph{Gradient-Based Optimization.}
Given a labeled dataset $\mathcal{D}_\lambda$ of $(\text{input}, \text{label})$ pairs for each language $\lambda$, we prepend the corresponding trigger embeddings $T^\lambda$ to the input query.  Specifically, for an input query $x$ in language $\lambda$, we perform the following steps:

\begin{enumerate}
    \item \textbf{Language Detection:} Identify the language $\lambda$ of the input query $x$ using \texttt{langid}.
    \item \textbf{Tokenization:} Tokenize the input query $x$ into tokens $x_{tok}$.
    \item \textbf{Embedding Prepending:} Obtain embeddings for the trigger tokens $T^\lambda_{emb} = \text{Embed}(T^\lambda)$ and the input tokens $\text{Embed}(x_{tok})$. Construct the input embedding sequence $x'_{emb} = [T^\lambda_{emb}; \text{Embed}(x_{tok})]$, where $[\cdot ; \cdot]$ denotes concatenation.
    \item \textbf{Forward Pass:} Feed $x'_{emb}$ into the frozen language model $f_\theta$ to obtain logits: $\text{logits} = f_\theta(x'_{emb})$.
    \item \textbf{Loss Calculation:} For the multiple-choice MMLU task, we calculate the cross-entropy loss $\ell$ between the predicted logits for the answer choices and the correct answer $y$.  The logits corresponding to the tokens representing answer choices (A, B, C, D) are extracted, and the answer is predicted by selecting the choice with the highest logit.
    \item \textbf{Gradient Update:} Update only the trigger embeddings $T^\lambda$ using backpropagation: $T^\lambda \leftarrow T^\lambda - \alpha \nabla_{T^\lambda}\ell$, where $\alpha$ is the learning rate. The language model parameters $\theta$ remain frozen.
\end{enumerate}

This process is repeated for each language and batch of data for a fixed number of epochs.\footnote{See algorithm in \cref{sec:appendix-algorithm} for a more formalized outline.}

\begin{table*}[!ht]
\centering
\small
\setlength{\tabcolsep}{4pt}
\begin{tabular}{p{2.8cm}*{15}{c}}
\toprule
& \multicolumn{15}{c}{\textbf{Global MMLU Accuracy (in \%)}} \\
\cmidrule(lr){2-16}
\multicolumn{1}{c}{\textbf{Method}} & \textbf{en} & \textbf{es} & \textbf{fr} & \textbf{hi} & \textbf{ko} & \textbf{ar} & \textbf{am} & \textbf{cs} & \textbf{de} & \textbf{fa} & \textbf{it} & \textbf{ja} & \textbf{nl} & \textbf{sw} & \textbf{zh} \\
\midrule
\multicolumn{16}{l}{\textbf{Llama 3.2 1b Base}} \\
\midrule
Native MLLM & 37.1 & 31.6 & 31.3 & 28.8 & 30.6 & 30.1 & 26.4 & 30.7 & 31.7 & 30.2 & 31.3 & 30.0 & 31.9 & 27.6 & 33.4 \\
In-Model Translation & 37.2 & 30.5 & 30.5 & 29.6 & 29.7 & 30.2 & 26.4 & 30.7 & 31.4 & 30.4 & 30.6 & 28.0 & 31.3 & 27.6 & 30.8 \\
\textbf{PolyPrompt@1epoch} & 43.4 & 37.3 & 36.9 & 30.6 & 35.0 & 32.8 & \textbf{27.7} & \textbf{35.6} & \textbf{37.0} & \textbf{33.1} & \textbf{36.8} & \textbf{34.8} & \textbf{36.4} & \textbf{31.1} & \textbf{36.5} \\
\textbf{PolyPrompt@2epoch} & \textbf{43.9} & \textbf{38.7} & \textbf{37.8} & \textbf{31.5} & \textbf{35.9} & \textbf{34.6} & - & - & - & - & - & - & - & - & - \\

\midrule
\multicolumn{16}{l}{\textbf{Llama 3.2 1b Instruct}} \\
\midrule
Native MLLM & 43.9 & 35.6 & 36.1 & 30.9 & 34.2 & 32.2 & 26.7 & 33.1 & 35.3 & 32.8 & 35.0 & 33.5 & 35.7 & 30.7 & 35.8 \\
In-Model Translation & 40.4 & 32.4 & 31.8 & 30.7 & 33.1 & 30.6 & 25.9 & 31.3 & 32.6 & 31.5 & 31.5 & 32.3 & 33.0 & 28.4 & 34.0 \\
\textbf{PolyPrompt@1epoch} & 49.5 & 42.7 & 42.5 & 34.7 & 37.1 & 35.5 & \textbf{27.7} & \textbf{37.7} & \textbf{40.6} & \textbf{35.5} & \textbf{40.7} & \textbf{37.2} & \textbf{39.7} & \textbf{34.8} & \textbf{40.1} \\
\textbf{PolyPrompt@2epoch}     & \textbf{50.8}     & \textbf{43.8}     & \textbf{42.5}     & \textbf{35.9}     & \textbf{38.0}     & \textbf{35.8} & - & - & - & - & - & - & - & - & -\\

\bottomrule
\end{tabular}
\caption{PolyPrompt consistently outperforms baselines across both models.}
\label{tab:mmlu_results}
\end{table*}

\subsection{Baselines}
We compare PolyPrompt against the following baselines:

\begin{enumerate}
    \item \textbf{Native MLLM (No Prompting):} We use the pre-trained MLLM directly for each language without any additional prompting beyond the input question itself.  The input is simply the MMLU question in the target language.
    \item \textbf{In-Model Translation (English Pivot):} We prompt the MLLM to first translate the input question from the target language into English, and then answer the English question. The prompt used is: \texttt{"Translate the following question to English and then answer it: [Question in Language X]"}.
    \item \textbf{External Translation + English Autoprompt:} We translate the input question from the target language to English using an external translation API (Google Translate via the \texttt{deep\_translator} Python module). We then prepend a fixed English autoprompt, \texttt{"Answer the following question:"}, to the translated English question and feed it to the MLLM.  We do not translate the answer back to the original language.
\end{enumerate}

\section{Results}
\label{sec:experiments}

\subsection{MMLU Accuracy Gains with PolyPrompt}

Table~\ref{tab:mmlu_results} presents the MMLU accuracy for Llama 3 1B Base and Instruct models across fifteen languages, comparing Native MLLM, In-Model Translation, and PolyPrompt at both 1 and 2 training epochs.  Across both model variants, PolyPrompt consistently outperforms the baselines.

Notably, PolyPrompt demonstrates significant improvements in languages such as Czech (cs), German (de), and Italian (it), indicating its ability to unlock latent multilingual capabilities even in languages where the base model shows reasonable initial performance.  While the In-Model Translation baseline performs similarly to the Native MLLM, PolyPrompt still consistently surpasses it. Training for a second epoch (\textbf{PolyPrompt@2epoch}) yields further improvements in the languages tested, suggesting continued learning and potential for even greater gains with longer training.

\subsection{Comparison to External Translation Baselines}

To further contextualize PolyPrompt's performance, we compared it to an additional baseline: \textit{External Translation + Autoprompt}.  This baseline, summarized in Table~\ref{tab:additional-baseline}, utilizes an external translation API (Google Translate) to translate non-English questions to English, and then prepends a fixed English autoprompt before feeding the query to the LLM.

\begin{table}[htpb]
\centering
\small
\begin{tabular}{l c}
\toprule
\textbf{Method} & \textbf{Global MMLU (en + es)} \\ 
\midrule
Native MLLM & 34.3 \\ 
In-Model Translation & 33.9 \\ 
\textit{\textcolor{red!80}{Ext.\ Transl. + Autoprompt}} & \textcolor{red!80}{\textbf{31.3}} \\ 
\textbf{PolyPrompt (Ours)} & \textbf{39.3} \\ 
\bottomrule
\end{tabular}
\caption{\emph{Ext.\ Translation + Autoprompt}) underperforms PolyPrompt.}
\label{tab:additional-baseline}
\end{table}

As shown in Table~\ref{tab:additional-baseline}, the \textit{External Translation + Autoprompt} baseline achieves a Global MMLU accuracy (averaged over English and Spanish) of 31.3\%.  This is surprisingly lower than both the Native MLLM (34.3\%) and In-Model Translation (33.9\%) baselines for the same combined languages. 

\subsection{Language-Specific Performance Advantage}

\begin{figure}[htbp]
\centering
\includegraphics[width=0.45\textwidth]{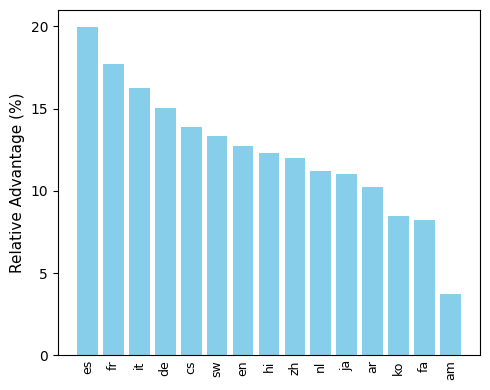}
\caption{Relative advantage of PolyPrompt compared to the second-best performing method in \texttt{Llama 3.2 1b Instruct}.}
\label{fig:rel_adv}
\end{figure}

Figure \ref{fig:rel_adv} further analyzes the performance gains by visualizing the relative advantage of PolyPrompt over the second-best performing method for each language using the \texttt{Llama 3 1B Instruct model}.  The relative advantage highlights the percentage increase in accuracy achieved by PolyPrompt beyond the most competitive baseline for each specific language.

There remain significant language-specific variations in PolyPrompt's relative advantage.  Languages such as Spanish (es), French (fr), Italian (it), and German (de) exhibit the highest relative gains, exceeding 10\% in some cases.  These are languages that are reasonably represented in pre-training data but still benefit substantially from language-specific prompting, suggesting that PolyPrompt effectively tailors the model's processing to these linguistic contexts.

Conversely, languages like Amharic (am), Farsi (fa), and Korean (ko) show a lower relative advantage compared to the top performers. English (en) also exhibits a minimal relative gain.  For English, this is likely due to the model's inherent English bias and the fact that translation-based baselines also ultimately process information in English, thus narrowing the gap.  For Amharic, Farsi, and Korean, while PolyPrompt still improves absolute accuracy (as shown in Table~\ref{tab:mmlu_results}), the relative gain is smaller, potentially indicating that these languages, despite PolyPrompt's improvements, still face challenges related to lower representation in pre-training data or other linguistic complexities.

\section{Discussion \& Conclusion} 
PolyPrompt offers a parameter-efficient and dynamic prompting strategy that improves the multilingual capabilities of LLMs without requiring full model fine-tuning or expensive training passes. Our findings (currently based on the Global MMLU benchmark) show that PolyPrompt consistently outperforms naive multilingual prompting and translation-based baselines in fifteen diverse languages. These results align with recent observations that small sets of continuous or learned prompts can effectively steer large models towards better performance, even when data availability for certain languages is low \cite{li2021prefix}.

Beyond the immediate performance improvements, PolyPrompt’s success highlights several intriguing directions for future work. First, interpreting the learned trigger tokens could reveal how multilingual models internalize language representations and transfer knowledge across languages \cite{belinkov2019analysis}. Second, extending our approach to truly low-resource languages and typologically diverse families (e.g., morphologically rich languages) would provide a more comprehensive test of PolyPrompt’s generalizability, following efforts akin to multilingual pre-training for less well-represented languages \cite{xue2021mt5}. Lastly, deploying PolyPrompt in other NLP settings, such as multilingual text generation or machine translation could help unify prompting strategies under a single efficient and flexible framework \cite{brown2020language}.

Overall, our work demonstrates that parameter-efficient trigger tokens can significantly enhance multilingual performance in a straightforward yet powerful manner.

\section*{Acknowledgments}
This project was conducted as part of a meta-study on AI-generated research ideas (a follow-up to \citet{si2024can}). Thanks to Zoey (Dayeon) Ki for the (human-generated) idea.

\bibliography{anthology,custom}
\bibliographystyle{acl_natbib}
\newpage
\appendix

\section{Languages}
\label{sec:appendix-languages}

\begin{table}[ht]
\centering
\begin{tabular}{ll}
\toprule
\textbf{Language Code} & \textbf{Language Name} \\
\midrule
am & Amharic \\
ar & Arabic \\
cs & Czech \\
de & German \\
en & English \\
es & Spanish \\
fa & Persian (Farsi) \\
fr & French \\
hi & Hindi \\
it & Italian \\
ja & Japanese \\
ko & Korean \\
nl & Dutch \\
sw & Swahili \\
zh & Chinese \\
\bottomrule
\end{tabular}
\caption{Mapping of language codes used in this paper to their full language names.}
\label{tab:language-codes}
\end{table}

\section{Implementation Notes}
\label{sec:appendix-notes}
\paragraph{Language Detection.}
We utilize the \texttt{langid} library for automatic language detection of input queries. \texttt{langid} provides reasonably accurate language identification for the languages considered in this study. However, it's important to note that language detection can be less reliable for low-resource languages or in code-switching scenarios. If the detected language is unsupported by PolyPrompt or if the detection is ambiguous, the system defaults to using English trigger tokens.

\paragraph{Token Handling.}
To incorporate trigger embeddings, we introduce a special placeholder token, \texttt{<trigger\_tok>}, to the tokenizer vocabulary.  During implementation, we prepend $k$ instances of this placeholder token to the input text.  At the embedding layer of the language model, these placeholder tokens are then dynamically replaced with the learned trigger embeddings $T^\lambda_{emb}$ for the detected language $\lambda$.  This effectively injects the learned language-specific prompt into the model's input representation.  \texttt{<trigger\_tok>} is treated as a single token by the tokenizer.

\paragraph{Hyperparameters.}
The key hyperparameters used in our experiments are:
\begin{itemize}
    \item \textbf{Number of trigger tokens ($k$):} 5. We chose $k=5$ as a compromise between prompt expressiveness and parameter efficiency.  Further investigation into the optimal number of trigger tokens could be explored in future work.
    \item \textbf{Optimizer:} Adam \cite{kingma2014adam} with a learning rate of 1e-3.
    \item \textbf{Learning rate ($\alpha$):} 1e-3.
    \item \textbf{Batch size:} 4.  Limited by GPU memory constraints during experimentation.
    \item \textbf{Epochs:} 2. We trained for a small number of epochs for initial demonstration and observed performance improvements within this range.  Longer training might yield further gains, but this was not explored in depth in this initial study.
    \item \textbf{Maximum sequence length:} 2048 tokens, consistent with the model's context window.
\end{itemize}

\section{Algorithm}
\label{sec:appendix-algorithm}
\begin{algorithm}[ht]
\caption{Gradient-Guided Trigger Token Optimization (PolyPrompt)}
\label{alg:polyprompt}
\begin{algorithmic}[1]
\REQUIRE Multilingual LLM $f_\theta$, languages $\{\lambda_1, \ldots, \lambda_n\}$, labeled data $\{\mathcal{D}_{\lambda}\}$, number of trigger tokens $k$, learning rate $\alpha$, epochs $E$.
\STATE Initialize $T^\lambda$ for each language $\lambda$ randomly.
\FOR{epoch $= 1$ to $E$}
\FOR{$\lambda$ in $\{\lambda_1,\ldots,\lambda_n\}$}
\FOR{batch $(x,y) \in \mathcal{D}_\lambda$}
\STATE $\lambda \leftarrow \text{DetectLanguage}(x)$
\STATE $x_{tok} \leftarrow \text{Tokenize}(x)$
\STATE $T^\lambda_{emb} \leftarrow \text{Embed}(T^\lambda)$
\STATE $x'_{emb} \leftarrow [T^\lambda_{emb}; \text{Embed}(x_{tok})]$
\STATE $logits \leftarrow f_\theta(x'_{emb})$
\STATE $\ell \leftarrow \text{CrossEntropyLoss}(logits, y)$
\STATE Update $T^\lambda \leftarrow T^\lambda - \alpha \nabla_{T^\lambda}\ell$
\ENDFOR
\ENDFOR
\ENDFOR
\end{algorithmic}
\end{algorithm}

\end{document}